\pdfoutput=1
\documentclass[10pt,twocolumn,letterpaper]{article}

\usepackage{tikz}
\usetikzlibrary{positioning,calc,snakes}
\usepackage{wacv}

\usepackage{times}
\usepackage{epsfig}
\usepackage{graphicx}
\usepackage{amsmath}
\usepackage{amssymb}
\usepackage[outercaption]{sidecap}

\usepackage{array}
\usepackage{subcaption}
\usepackage{amsmath}
\usepackage[normalem]{ulem}

\newcommand{\CNNo}{$\text{CNN}_{\text{0}}$ }
\newcommand{\CNNw}{$\text{CNN}_{\text{w}}$ }
\newcommand{\Fo}{$\mathbf{F}_{\text{0}}$ }
\newcommand{\Fw}{$\mathbf{F}_{\text{w}}$ }
\newcommand{\RGBN}{\text{RGB}_\text{90}}
\newcommand{\RGBZ}{\text{RGB}_\text{0}}
\newcommand{\Rou}{\text{R}}

\usepackage[pagebackref=true,breaklinks=true,letterpaper=true,colorlinks,bookmarks=false]{hyperref}

\wacvfinalcopy 

\def\wacvPaperID{467} 

\ifwacvfinal\fi
\begin{document}

\title{BRDF Estimation of Complex Materials with Nested Learning}

\author{Raquel Vidaurre \hspace{0.5cm} Dan Casas \hspace{0.5cm} Elena Garces \hspace{0.5cm} Jorge Lopez-Moreno\\
URJC, Madrid \hspace{3.5cm} Desilico Labs\\
{\tt\small \{raquel.vidaurre|dan.casas\}@urjc.es  \{elena.garces|jorge.lopez\}@desilico.tech }
}

\twocolumn[{
	\renewcommand\twocolumn[1][]{#1}
	\maketitle
	{
		\centering
		\vspace{-0.3cm}
		\includegraphics[width=\linewidth,trim={0 0 0 20cm},clip]{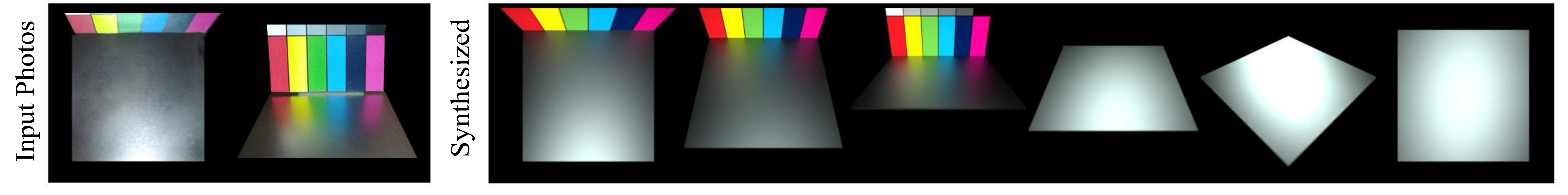}
		\captionof{figure}{
			From just two input images (left) our method is capable of estimating the BRDF parameters (right, synthesized from novel viewpoints) of complex materials.
			\label{fig:teaser}
		}
	}
	\vspace{0.5cm}
}]

\thispagestyle{empty}

%
%

\begin{abstract}
   The estimation of the optical properties of a material from RGB-images is an important but extremely ill-posed problem in Computer Graphics. While recent works have successfully approached this problem even from just a single photograph, significant simplifications of the material model are assumed, limiting the usability of such methods.
   The detection of complex material properties such as anisotropy or Fresnel effect remains an unsolved challenge. 
   We propose a novel method that predicts the model parameters of an artist-friendly, physically-based BRDF, from only two low-resolution shots of the material. Thanks to a novel combination of deep neural networks in a nested architecture, we are able to handle the ambiguities given by the non-orthogonality and non-convexity of the parameter space. To train the network, we generate a novel dataset of physically-based synthetic images.
   We prove that our model can recover new properties like anisotropy, index of refraction and a second reflectance color, for materials that have tinted specular reflections or whose albedo changes at glancing angles.  
\end{abstract}

	\section{Introduction}
	\label{sec:introduction}
	%
	One of the main problems in Computer Graphics is the creation of synthetic images which are indistinguishable from real world objects, a process known as rendering. Among the existing rendering techniques, physically-based rendering algorithms achieve plausible results by computing the way light bounces through the scene, an operation in which the right definition of a material is critical to convey realistic results.
	%
	%
	To describe and encapsulate the material reflectance behavior, a common model is the a Bidirectional Reflectance Distribution Function (BRDF), but the acquisition and measurement of the BRDF of real materials is a big challenge due to the sheer number of uncontrolled optical phenomena affecting the light transport on the material surface. While there are specific techniques for this purpose~\cite{filip2013brdf,filip2014effective,Ngan2005,filip2014template,vavra2016minimal}, it is a slow, constrained and expensive process. 
	A usual approach to circumvent this problem is to use a simplified analytic model controlled by a set of parameters~\cite{Lawrence:2006:IST,weidlich2007arbitrarily,dupuy2015extracting,Brady:2014}. Despite the expressiveness of these models, when trying to reproduce a real world material, this process of tuning parameters by hand is a time consuming and error-prone task for digital artists.
	
\begin{figure}[tb]
		\centering
		\includegraphics[width=0.6\columnwidth]{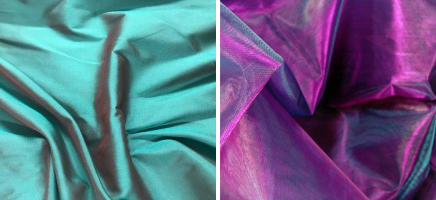}
		\caption{Examples of anisotropic materials}
		\label{fig:anisotropic}
	\end{figure}

	Anisotropic models are capable of depicting a great variety of materials (Figure~\ref{fig:anisotropic}), however, choosing the appropriate model for the appearance is critical, and, sometimes, the flexibility required by an artist to tweak the parameters is not aligned with the facility of finding them in an optimization procedure. The approach of Aittala et al.~\cite{aittala2015twoshot} fits a learned model from Brady et al.~\cite{Brady:2014} by means of a flash-no flash image pair and using non-linear optimization methods. 
	In opposition to an empirical BRDF, in this work we chose a physically-based, industry-oriented model~\cite{maxwell} widely adopted in the film and video game industries. Designed with artistic expressiveness as main goal, it has the key challenge of having a non-orthogonal, non-convex parameter space, making existing optimization methods unsuitable.
	
	
	In this work we propose a novel real-time method based on neural networks to automatically estimate the BRDF parameters of an arbitrary material using just two low resolution images. We estimate ten parameters that model complex reflectance effects, including anisotropy. As this is a highly ill-posed problem, we make some assumptions: we assume that pictures are taken in a controlled (but low cost) environment and, we also assume planar objects with homogeneous materials.  
	
	Recent methods based on deep learning techniques have proved to be successful in fitting the parameters of isotropic material models from arbitrary single images~\cite{meka2018lime}. However, real world materials like brushed metals or cloth cannot be modelled with simple isotropic models like Phong~\cite{lafortune1994using} or Ward~\cite{ward1992measuring}.
	
	This paper makes the following contributions:
	\begin{itemize}
		\item A method that estimates the parameters of an anisotropic BRDF from two shots of the material at 30$^{\circ}$ and 90$^{\circ}$ degrees.
		\item A hybrid deep learning architecture that combines feature learning with nested fully-connected layers that leverage dependencies in the parameter space of the model.
		\item A novel loss function that combines pure classification with gaussian smoothing to perform regression of the parameters. 
		\item A novel synthetic dataset of planar samples of anisotropic materials rendered under multiple viewpoints.
	\end{itemize}


	%
	%
	%
	%
	\section{Related Work}
	\label{sec:related_work}
	
	\paragraph*{Anisotropic Materials} The final appearance of anisotropic materials is determined by multiple factors including the mechanical properties of the material along with the viewer and light source orientation. Therefore, accurately capturing these complex interactions require the use of expensive gonioreflectometers~\cite{filip2013brdf,filip2014effective}, or controlled setups with various sampling strategies~\cite{Ngan2005,filip2014template,vavra2016minimal}.
	
	Similarly, a few parametric models exist to simulate anisotropic behaviors based on microfacets~\cite{weidlich2007arbitrarily,dupuy2015extracting}, Wasserstein coordinates~\cite{bonneel:wasserstein}, or non-physically-based models with editable capabilities~\cite{Lawrence:2006:IST}. Among them, the closest to our work is the method of Aittala et al.~\cite{aittala2015twoshot}. 
	Equally relying on an image pair, they use Levenberg-Marquardt to optimize the orthogonal parameters of a machine-learned microfacets model~\cite{Brady:2014}. 
	In contrast, our anisotropic model is artist-friendly, meaning in practice that different set of parameters might yield the same appearance, and thus, making the problem unsuitable for conventional optimization methods.
	In this work, we handle this ambiguity by modeling the parameter relationships with a nested neural network that estimates the parameters of our model, given only two pictures of the material taken under two different orientations.
	
	
	Other methods that model and capture anisotropy cannot be used in a wild and uncontrolled environment. We refer to the state-of-the-art paper of Guarnera et al.~\cite{guarnera2016brdf} for a complete analysis of anisotropic models and acquisition methods. We present the first low cost and easy to use a method that capture anisotropic materials using a highly expressive and artist-friendly model.
	
	
	\begin{table*}
		\small
		\begin{center}
			\begin{tabular}[c]{ | m{1.8cm} | m{2.3cm}| m{0.8cm} | m{3cm}| m{7cm} |}
				\hline
				Method &  Arquitecture(s) & Loss &  Material Model & Output \\ \hline
				\cite{innamorati2017decomposing} & U-Net & \textit{rloss} 
				& Phong~\cite{lafortune1994using}& Images: albedo, irradiance, speculars, and occlusions \\ \hline
				\cite{li2017modeling} &  [1] CNN+FC; [2] U-Net &  \textit{rloss} 
				& SVBRDF  Ward~\cite{ward1992measuring} & [1] RGB specular and roughness (4 params); [2] Images: albedo, normals \\ \hline
				\cite{yu2017pvnn} &  CNN\footnotesize{Resnet}\normalsize+FC & \textit{rloss} 
				& Nielsen~\protect\cite{Nielsen2015} & 15 parameters \\ \hline    
				\cite{georgoulis2017reflectance} & [1] U-Net; [2] CNN+FC, U-Net & \textit{rloss}  
				&  [1] R.~Map~\cite{horn1979calculating}  [2] Phong~\cite{phong1975illumination} & [1] image; [2] 7 params + image spherical illumination \\ \hline   
				\cite{liu2017material} &   CNN + FC & \textit{rloss, ploss} 
				& Directional Statistics BRDF \cite{lombardi2012reflectance} & normals, environment map, 108 material params  \\ \hline
				\cite{meka2018lime} &  Stacked U-Net & \textit{rloss, ploss} 
				& Blinn-phong \cite{blinn1977models} & 4 params (RGB albedo and specular coefficient)  \\ \hline
				\cite{kim2017lightweight} &  CNN/Voxel  & \textit{ploss} 
				& Isotropic Ward \cite{ward1992measuring} & 5 parameters \\ \hline    
				\textbf{Ours} & CNN+nestedFC  & \textit{ploss} 
				& Maxwell~(Ashikhmin-Shirley's)~\cite{Ashikhmin2000} & 10 parameters \\ \hline
			\end{tabular}
		\end{center}
		\caption{Comparison of deep learning-based methods for isotropic materials. All the methods take as input a single image except the method of Kim et al.~\protect\cite{kim2017lightweight} that needs RGBD images, the method of Li et al.~\cite{li2017modeling} which constrains the input to planar samples of metal, wood or plastic, and our method, that requires two captures. Note that our method is the first to fit a complex anisotropic model, while all the other methods use isotropic material models.} 
		\label{table:rwisotropic}
	\end{table*}

	\paragraph*{Isotropic Materials} On the other hand, isotropic materials, have received much more attention due to their simplicity to capture~\cite{Matusik:2003}, model and render in virtual 
	environments~\cite{xu2016minimal}. Non deep-learning methods for isotropic models include: capturing SVBRDF and normal maps from two shots~\cite{aittala2015twoshot}, single shots~\cite{aittala2013practical}, dictionary-based~\cite{hui2017reflectance}, or based on neural texture synthesis~\cite{aittala2016reflectance}. Others take as input RGBD images like \cite{wu2016simultaneous} or \cite{lombardi2016radiometric}.
	
	However, recent works that rely on deep learning techniques have proved to outperform conventional optimization. 
	We can categorized these methods in two big groups according to whether the loss function is computed as: 1) the pixel-wise difference between the ground truth image and a rendered image computed with the parameters predicted by the model, i.e. the render loss (\textit{rloss}) \cite{innamorati2017decomposing,li2017modeling,yu2017pvnn,georgoulis2017reflectance,meka2018lime}; 2) the difference between the ground truth and the estimated parameters, i.e. the parameter loss (\textit{ploss}) \cite{kim2017lightweight,liu2017material,meka2018lime}. Methods that rely on the \textit{rloss} are in general more precise as the training learns how much a variation of parameters affect the desired results. 
	However, they need the render function to be computed almost in real time. Our anisotropic model is computationally complex and takes several minutes to compute, making our network unfeasible to be trained with the \textit{rloss}. We get inspiration from the work of Kim et al.~\cite{kim2017lightweight}, that takes as input an RGBD image, and uses the \textit{ploss} enhanced with color histogram-based regularizators to find the five parameters of an isotropic Ward model~\cite{ward1992measuring}. On the contrary, we use a nested architecture designed to reduce the ambiguity given by the non-orthogonality of the parameters. With the exception of Meka et al.~\cite{meka2018lime}, the other methods estimate the material parameters by regression. Instead, we pose the problem as a classification task which results in more robust estimations with greater explicability.
	Table~\ref{table:rwisotropic} shows a summary of the deep learning based methods for isotropic models.
	\paragraph*{Datasets for learning-based methods.} Any supervised machine learning-based method requires an annotated dataset for training. Given the variety of models, each method has developed its own dataset specific for the task at hand without publishing it, with a few exceptions. The dataset of OpenSurfaces~\cite{bell2013opensurfaces} with perceptual Ward~\cite{pellacini2000toward} parameters obtained via crowdsourcing, SynBRDF~\cite{kim2017lightweight} containing 500K RGBD images rendered under a variety of natural illumination (5000 materials, 5000 shapes), Objects under Natural Illumination Database~\cite{lombardi2012reflectance} with six objects taken under five natural illumination environments with calibrated ground-truth geometry and illumination. These datasets rely on simple isotropic models of materials. Our MaxBRDF dataset, which we plan to make available, provides a very complex set of planar samples of anisotropic materials.


	
	%
	\section{BRDF Model Analysis}   
	\label{sec:bsdf_model}
	
	
	We focused our work in the standard BSDF of the commercial software Maxwell Render~\cite{maxwell}, the reason is twofold: first, Maxwell Render is an unbiased physically based rendering engine, capable of generating realistic synthetic images reproducing a real setup~\cite{yu2017pvnn}. And second, Maxwell's underlying BSDF is designed for artists, with very expressive but non-convex and non-orthogonal parameters: this is a challenging scenario for classic optimization approaches, but a nice testbed to explore the potential of CNN architectures for parameter fitting.
	
	%

\begin{figure}
		\centering
		\includegraphics[width=0.5\columnwidth]{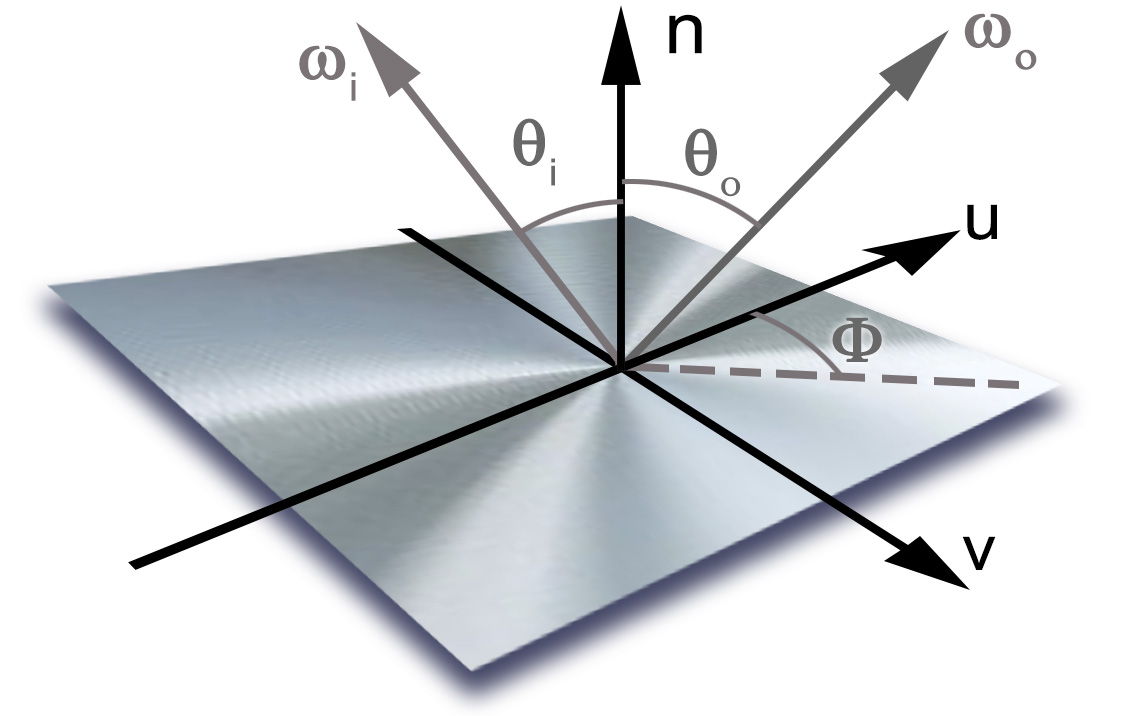}
		\caption{Diagram of Maxwell Render's BRDF model. The vectors $\omega_i$ and $\omega_o$ are generally associated to incoming light and viewer direction.}
		\label{fig:bsdf_diag}
	\end{figure}

	Maxwell Render's underlying BSDF is a reflectance (BRDF) and transmittance (BTDF) model defined by twenty-five (25) parameters. 
	Although artists usually combine this model with spatially varying (SV) textures, often layered in a stack of multiple BSDFs, for the purpose of this work we match the material in isolation, although it could be extended to SVBRDF with further statistical analysis, such as the texture transfer technique described by Aittala \textit{et al}.~\cite{aittala2015twoshot}. We focus our study in the reflectance component (BRDF) of the BSDF, as shown in Figure~\ref{fig:bsdf_diag} and defined by the equations described next.
	
	The BRDF defining the opaque reflectance is an adaptation of Ashikhmin-Shirley's model \cite{Ashikhmin2000} able to represent complex behaviors like anisotropy and Fresnel reflections. The main  difference is the elimination of the original diffuse term $\rho_d$, and the inclusion of a 90$^{\circ}$-reflectance value which is dominant at glancing angles and the reversion to a purely diffuse model when the roughness is approaching 100\%, as shown in Equation \ref{eqn:BRDF1} which defines the BRDF model:
	
	\begin{eqnarray}
	\left.\begin{aligned}
	f(\omega_i,\omega_o) &= 
	\begin{cases}
	f_m(\omega_i,\omega_o) & \Rou \leq 90  \\
	\alpha \text{ } f_m(\omega_i,\omega_o) + \\(1-\alpha) \RGBZ \cos\theta_o  & \Rou > 90   
	\end{cases} \\
	\alpha &= 0.1 |100-\Rou|
	\end{aligned}\right.
	\label{eqn:BRDF1}
	\end{eqnarray}
	
	\begin{eqnarray} 
	\left.\begin{aligned}
	f_m(\omega_i)&=  (\RGBN \text{ }(1-\cos\theta_o)\text{ } + \\ &\RGBZ \cos\theta_o\text{ }) \text{ }\rho_s(\omega_i,\omega_o) \text{ } f_r(\omega_i) \\
	\end{aligned}\right.
	\label{eqn:BRDF2}
	\end{eqnarray}

	\begin{eqnarray} 
	\left.\begin{aligned}
	f_r(\omega_i)&= 
	\frac{1}{2} \bigg(\frac{k^2\cos^2\theta_i+(N_d\cos\theta_i-1)^2}{k^2\cos^2\theta_i+(N_d\cos\theta_i+1)^2} \\& + \frac{k^2+(N_d-\cos\theta_i)^2}{k^2+(N_d+\cos\theta_i)^2} \bigg) 
	\end{aligned}\right.
	\label{eqn:BRDF3}
	\end{eqnarray}
	
	\begin{eqnarray} 
	\left.\begin{aligned}
	\rho_s(\omega_i,\omega_o)&= 
	\frac{\sqrt{(n_u+1)+(n_v+1)}}
	{8\pi} \\& \cdot \frac{(n\cdot h)^{n_u \cos^2\phi+n_v\sin^2\phi}}{(h\cdot k)\max((n\cdot\omega_i),(n\cdot\omega_o))}\\
	\end{aligned}\right.
	\label{eqn:BRDF4}
	\end{eqnarray}
	
	Where $f_r$ is the Fresnel component, and $\rho_s$ is the specular component (See equations \ref{eqn:BRDF3} and \ref{eqn:BRDF4}). Please note that the operations with $\RGBZ, \RGBN$ are not representing a simple multiplication of RGB channels, as the luminance is controlled by the Fresnel component. Only the chromatic hue is conserved, yielding an interesting feature for artists to represent tinted Fresnel reflections in materials such as metals.
	
	Based on these equations, a wide range of materials can be represented by artists tuning the following ten parameters:
	
	\begin{itemize}
		\item \textbf{$\theta_0$: Roughness ($\Rou$)} determines the tiny imperfections that make a material reflect the light in a diffuse way. Its influence on the final appearance is very significant and most of the remaining parameters depend directly on its value. We can observe it for $\RGBZ$ and $\RGBN$ in Equation \ref{eqn:BRDF1} and \ref{eqn:BRDF2}, for \textbf{anisotropy} in equation \ref{eqn:BRDF5} and for IOR (N$_d$) in Equation \ref{eqn:BRDF2}, as the term $\rho_s$ is multiplying the Fresnel term ($f_r(\omega_i)$).
		\item \textbf{$\theta_{1-2}$: Anisotropy and Anisotropy Angle} address to how directional the light is reflected. Anisotropy is a common property in brushed metals.  The $n_u$ and $n_v$ exponents depend on the value of $\Rou$ and \textbf{anisotropy}, and control the shape of the specular lobe in a similar fashion to the Ashikhmin-Shirley's model \cite{Ashikhmin2000} as defined in Equation~\ref{eqn:BRDF5}. The \textbf{anisotropy angle} controls the direction of the effect, by setting the origin w.r.t. the $\Phi$ angle within the plane defined by the surface U,V vectors.
		\begin{eqnarray} 
		\left.\begin{aligned}
		n_u &= (0.001 + 0.00999 \Rou)^{-3}\\
		n_v &= (n_u^{-1/3} + Anisotropy^3)^{-1}
		\end{aligned}\right.
		\label{eqn:BRDF5}
		\end{eqnarray}

		\item \textbf{$\theta_{3}$: N$_d$} is the Index Of Refraction (IOR). It determines how light propagates through the material and the amount of reflection. 
		A material with N$_d$ set to a low value will have weak reflections, while high values will create a mirror-like effect. The complex part of the IOR (the refraction extinction coefficient $k$) is set to zero in our data, due to its negligible effect in opaque materials.        
		\item $\theta_{4-6}$: \textbf{Reflectance 0}$^{\circ}$ ($\RGBZ$) refers to the color of a material, i.e. how light is reflected when it hits the surface and is seen from a front view. It is composed of 3 parameters, the reflectance color in RGB space. 
		\item $\theta_{7-9}$: \textbf{Reflectance 90}$^{\circ}$ ($\RGBN$), like the previous one, its three parameters determine the amount of light that is reflected, but in this case, when it is seen at 90$^{\circ}$. The influence of each of the reflectance colors depends strongly on the roughness value $\Rou$, but also on the N$_d$ value. In general, the $\RGBZ$ value is more prevalent than $\RGBN$ on the final appearance, due to the form factor ($\cos\theta_o$) which adds weight to the camera-facing surface pixels and the use of a Lambert model for high values of R.

	\end{itemize}
	
	The equations show no interdependencies between Anisotropy and IOR(Nd), but in our analysis of the material appearance variation produced by fixing one parameter and varying the other for regular increments of R, the anisotropy-dependent term $\rho_s(\omega_i,\omega_o)$ produced a greater variation per pixel for a given camera response. Intuitively, for increasing values of $N_d$ the existing reflections on the surface vary proportionally in luminance and color, while an equivalent differential variation on anisotropy, affects the shape of the reflection lobe $\rho_s$, drastically altering the shape and position of the reflections over the surface (see supplementary material).

	\paragraph*{Optimization Strategy} 
	Any attempt to fit this set of ten non-convex parameters will face the problem of local minima produced by first, multiple possible combinations producing the same spatially local values, and second by overlaid appearance effects masking each other. In an ideal case we would try to detect and isolate the effect of one parameter over the others at a given pixel, but with traditional optimization approaches, this has proven to be challenging.

	On the early stage of the research, we tried to fit the three parameters of $\RGBZ$ albedo using a classical optimization method, but the results were limited. These preliminary results and the nature of our parameter space, led us towards deep neural networks, more suitable for such scenarios, materializing the architecture described in the next section.
	%
	\section{BRDF Estimation}\label{sec:regression}
	
\begin{figure}
		\centering
		\includegraphics[width=\columnwidth]{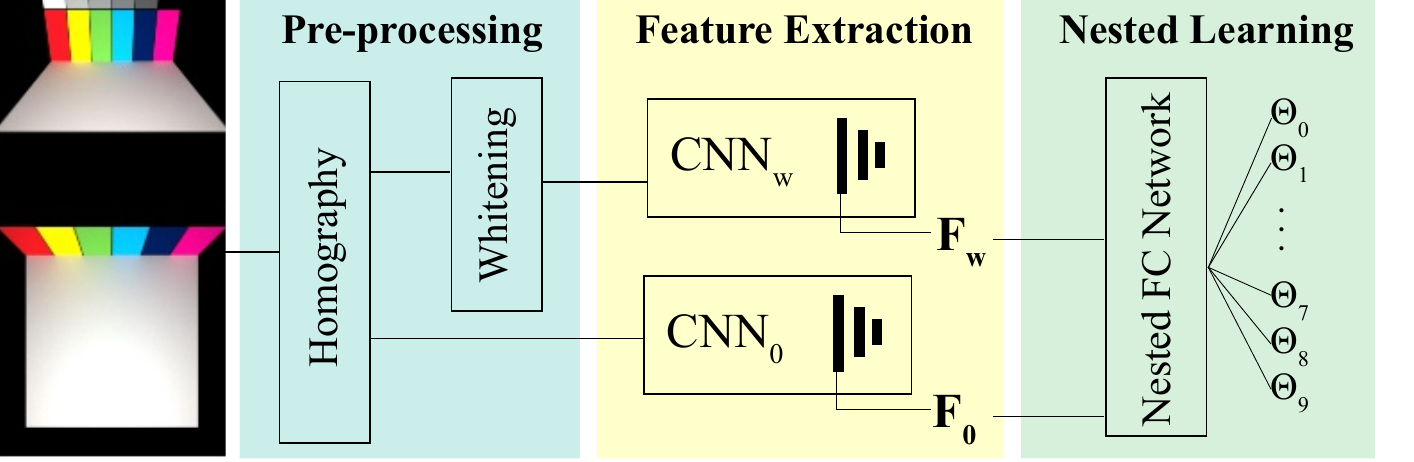}
		\caption{Pipeline of our method.}
		\label{fig:pipeline}
	\end{figure}

	Our goal is to provide a low-cost tool to easily capture the appearance of complex anisotropic materials. To this end, 
	we train a combination of deep neural networks that, taking as input two shots of the target material, yields the values of the ten parameters of our anisotropic model described in Section \ref{sec:bsdf_model}. 
	Figure~\ref{fig:pipeline} shows the pipeline of our method. 
	First, an initial pre-processing step transforms the input pair into a more suitable representation by means of computing its planar homography and whitening the images. The \textit{homographied} images are fed into two CNNs that extract a set of features, which are the input to our nested neural network that predicts the material parameters.
	
	\begin{figure}
		\centering
		\begin{subfigure}[t]{0.24\columnwidth}
			\includegraphics[width=\textwidth]{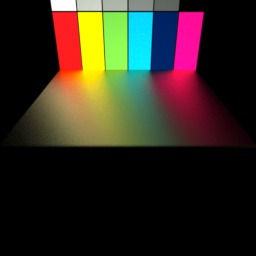}
			\caption*{\scriptsize{Rendered 30$^{\circ}$}}
		\end{subfigure}
		\begin{subfigure}[t]{0.24\columnwidth}
			\includegraphics[width=\textwidth]{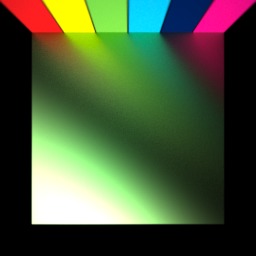}
			\caption*{\scriptsize{Rendered 90$^{\circ}$}}
		\end{subfigure}
		\begin{subfigure}[t]{0.24\columnwidth}
			\includegraphics[width=\textwidth]{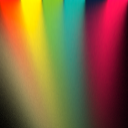}
			\caption*{\scriptsize{Homographied 30$^{\circ}$}}
		\end{subfigure}
		\begin{subfigure}[t]{0.24\columnwidth}
			\includegraphics[width=\textwidth]{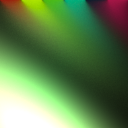}
			\caption*{\scriptsize
				{Homographied 90$^{\circ}$}}
		\end{subfigure}
		
		\caption{Two of the 14 views of a sample of our MaxBRDF dataset, and their corresponding \textit{homographied} version after the pre-processing step described in Section \ref{sec:preprocessing}}
		\label{fig:synthetic_dataset}
	\end{figure}
	
	\subsection{Pre-processing}\label{sec:preprocessing}
	
	Our input consists of two views of the material captured at 30$^{\circ}$ and 90$^{\circ}$, as we show in Section \ref{sec:dataset}, we empirically observed that these two views alone summarize most of the interesting features of the material.
	%
	%
	Before feeding the input image pair to the network, we perform a pre-processing step. We leverage the constraint to planar materials and find the homography that warps each input material to the orthogonal camera plane, resulting in a squared image which we referred to as \textit{homographied} input. This step facilitates the training and inference of the network particularly in real images that may have slightly different backgrounds. 
	This transformation is shown in figures \ref{fig:synthetic_dataset} and \ref{fig:capture_setup} for synthetic and real data, respectively. 
	Additionally, the images are whitened to remove color correlations which hinder the estimation of parameters such as \textit{roughness} and the index of refraction $N_d$. 
	
	\subsection{Parameter Loss}
	Before defining each of the modules of our pipeline, we describe the loss function that our network uses.
	Our goal is to predict the values for a set of parameters, a problem typically solved as a regression task~\cite{kim2017lightweight,li2017modeling}. However, by posing the problem as a classification task~\cite{meka2018lime}, we reduce the complexity: from having to find a number within the whole domain of the real numbers, to a fixed set of unknown values.
	We empirically found that discretizing each parameter into 100 uniform bins provides enough accuracy and convergence of the models.
	In contrast to most of the previous methods, which encode the material albedo using RGB space \cite{meka2018lime}, we leverage the \textit{perceptually uniform} distribution of CIELAB space and encode the $\RGBZ$ and $\RGBN$ parameters in LAB.
	This enforces that a misclassified bin in this color space directly match human perception, improving the accuracy of the predictions, as we show in Section \ref{sec:evaluation_and_results}.

\begin{figure}
		\centering
		\includegraphics[width=0.7\columnwidth]{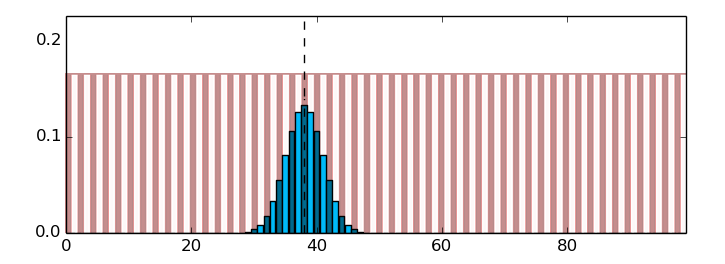}
		\caption{Input label after Gaussian smoothing (blueish bins), originally a one-hot value of 38 (dashed line).}
		\label{fig:gauss}
	\end{figure}

	A vanilla classification loss function computes the cross-entropy loss per each parameter $\theta_i$ as
	\begin{equation}
	\mathcal{L}_{\theta_{i}}(X,Y) = - \sum_{j=1}^{N} y_j \log (x_j),
	\label{eq:vanilla_loss}
	\end{equation}
	where $N=100$ is the number of bins/classes, $x_j$ is the probability for the class, and $y_j$ is the true label codified using one-hot encoding. 
	%
	%
	%
	%
	%
	%
	%
	However, in our particular scenario of regression-by-classification, a weakness of the loss defined in Equation \ref{eq:vanilla_loss} is that \textit{almost} correctly classified parameters are dismissed. 
	%
	In other words, vanilla classification loss makes sense when there is no relationship between the classes, however, in our case, small variations in the parameters produce imperceptible variations in the render image. Therefore, mispredicted classes located close by in the parameter space should be rewarded.
	To this end, we apply a Gaussian 1-dimensional discrete filter of radius nine to the one-hot encoded labels $Y$ (see Figure \ref{fig:gauss}) and redefine our loss as:
	\begin{equation}\label{eq:ploss}
	\hat{\mathcal{L}}_\theta(X,\hat{Y}) = - \sum_{j=1}^{N} \hat{y}_j \log (x_j)
	\end{equation}
	where $\hat{y}_j$ is the value after application of the Gaussian smoothing to the whole vector $Y$.
	This change led to smooth results and patently improved the accuracy of our model.
	
	%
	%
	%
	\subsection{Feature Extraction}
	\label{sec:feature_extraction}

	\begin{figure}
		\centering
		\includegraphics[width=\columnwidth]{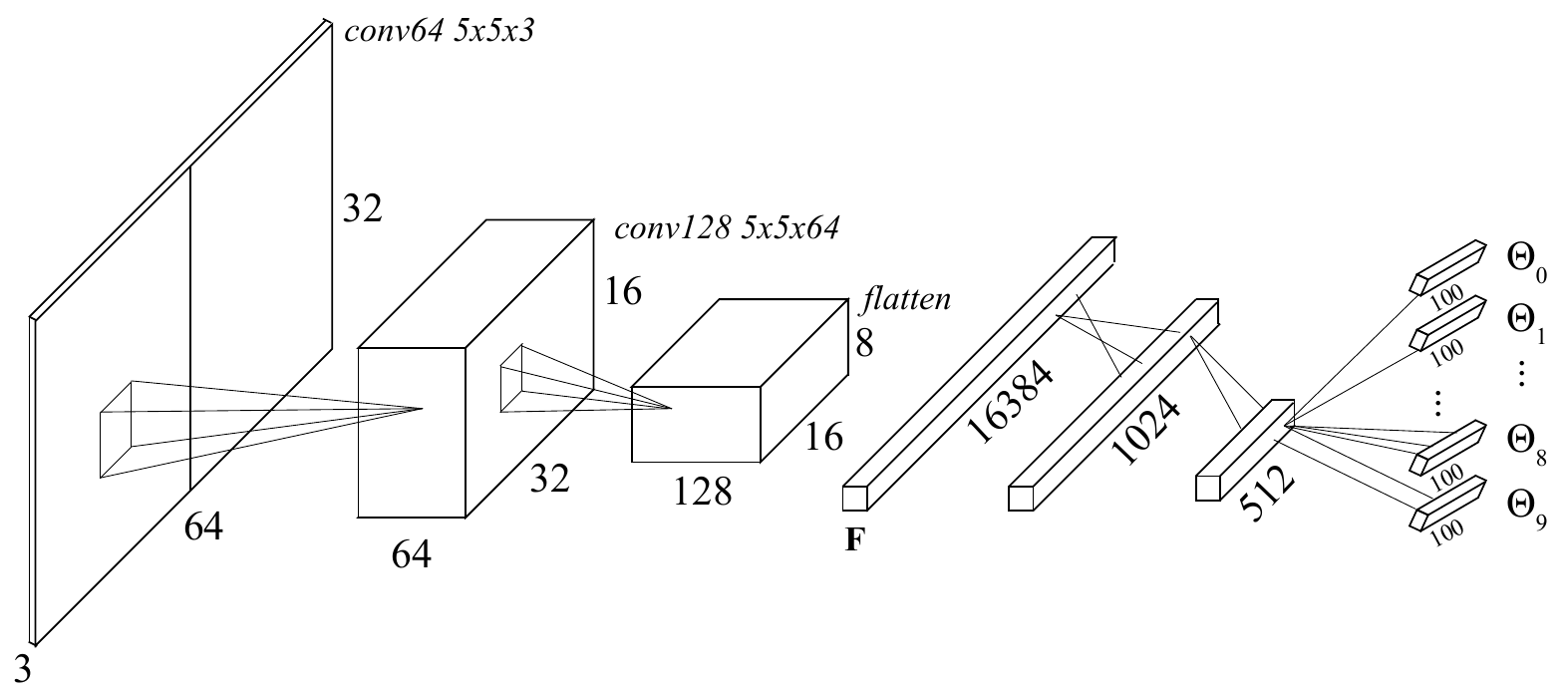}
		\caption{Architecture of each of the two CNN of our Feature Extraction module. We obtain the feature vectors \Fo and \Fw corresponding to the output of the second convolutional layer for a network trained with the original data \CNNo and a network trained with the whitened data \CNNw}
		\label{fig:cnn}
	\end{figure}
	The first module of our pipeline aims at extracting highly discriminative features from the input pair.
	It consists of two identical convolutional layers with 64 and 128 5x5 filters, respectively. Each layer is followed by a ReLU (Rectified Linear Unit) activation, a batch normalization, an LRN layer (Local Response Normalization) and a max-pooling layer with a stride of 2. After that, we apply three dense (or fully connected) layers, all of them followed, again, by a ReLU layer. Between all the layers, we apply dropout (with 75\% of keeping values) to make the network more robust and avoid overfitting. A diagram of the network is shown in Figure~\ref{fig:cnn}.
	Note that, in this step, all the ten classifiers are part of the same network that learns shared weights in the hidden layers and diverges in the last fully connected layer. Such structure is usually known as multitask learning and it improves the accuracy, as the learning of similar but different tasks can help the network to recognize relations and adds some noise that helps generalization \cite{caruana1998multitask}.
	
	The total loss to train this network is given by the sum of the local losses (Equation~\ref{eq:ploss}) per parameter as follows:
	\begin{equation}
	\mathcal{L}_{total} = \frac{1}{P}\sum_{i=1}^{P} \hat{\mathcal{L}}_{\theta_i},
	\label{eq:global_loss}
	\end{equation}
	where $P$ is the number of parameters. 
	
	We train two identical end-to-end networks: the \CNNo network, which takes the original homographied pair as input; and the \CNNw network, which inputs are additionally whitened.
	The performance of these two networks is reasonable, as our ablative study shows in Section \ref{sec:evaluation_and_results}, but still far from the ground truth.
	The reason is that the dependencies between the parameters are implicitly learned by the network, increasing the difficulty of the learning process.
	In light of these results, and as we know the parameter dependencies from the model definition in Section \ref{sec:bsdf_model},
	we manually designed a nested fully-connected architecture that replaces the FC layers of the current CNN and combines the knowledge learned from both models, \CNNo and \CNNw.



	\subsection{Nested Learning}
	%
	Given the feature maps \Fw and \Fo obtained in the Feature Extraction module, we train a nested fully connected architecture hand-crafted specifically for the task of estimating the parameters of our reflectance model.
	Each fully connected block (FC) has the same structure as the fully connected block of the CNN of the Feature Extraction module, i.e. three layers with intermediate units of sizes 1024 and 512 with a softmax activation unit per parameter of size 100. Each FC layer is trained independently and differ in the input and the output as described below.
	
\begin{figure}
		\centering
		\includegraphics[width=0.7\columnwidth]{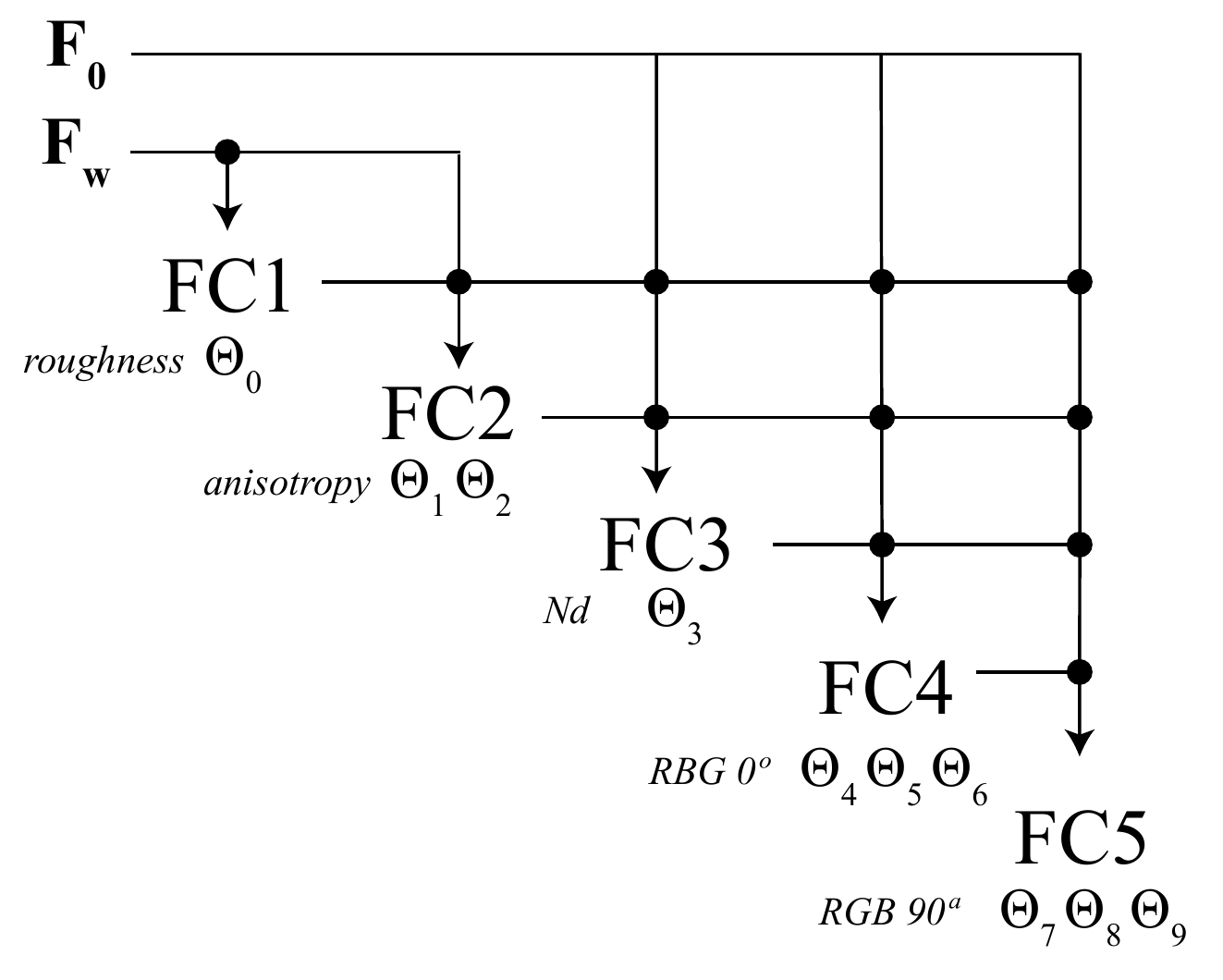}
		\caption{Nested Architecture. Each  $\bullet$ represents a concatenation of features. Each FC contains three fully connected layers with the same structure as the CNN baseline architecture for feature extraction. Note that the feature vector \Fw is used as input for FC1 and FC2, and \Fo feeds FC3, FC4 and FC5.}
		\label{fig:nested}
	\end{figure}

	From the definition of our BRDF model, in Equation \ref{eqn:BRDF1}, we observe that there is a dependency between the parameters, i.e. some parameters overwrite others.
	We explicitly model this dependency in the way feature maps and outputs of the FC layers are concatenated. The connections are depicted in Figure~\ref{fig:nested}, where the dots mean a concatenation operation between vectors that intersect. For example, the roughness parameter $\theta_0$ takes as input the feature vector \Fw and outputs 100 units; these units are therefore concatenated with the feature vector \Fw to form the input of the layer FC2. FC2 will learn two parameters $\theta_1$ and  $\theta_2$, which are again concatenated with the first parameter $\theta_0$ and the feature vector \Fo to form the input of the layer FC3. 
	
	
	\section{Evaluation and Results}
	\label{sec:evaluation_and_results}

	\subsection{Dataset and Training}
	\label{sec:dataset}

	%
	\paragraph*{MaxBRDF Dataset}
	It is well known that training deep learning methods require large amount of annotated data, however, annotating real world materials with their corresponding parameters of such complex model is unfeasible and error-prone even for skilled artists.
	%
	We therefore present a new physically-based rendered dataset, the MaxBRDF dataset, with much richer annotations than existing works, which we use to train our networks.
	
	Samples of the dataset are shown in supplementary material. We use Maxwell Render to generate 126K planar samples of 9K different materials varying our ten target parameters described in Section~\ref{sec:bsdf_model}. For each material, we render 14 different views. First, with the light source set at 45$^{\circ}$ of elevation above the sample, four rotations of the material with camera at 45$^{\circ}$, plus five orientations of the camera (15$^{\circ}$, 30$^{\circ}$, 45$^{\circ}$, 60$^{\circ}$ and 90$^{\circ}$), and finally, five additional views with the same camera directions but the light source at 90$^{\circ}$. In the former, we add a color checker, as the way the material reflects this colors can help detect some important features. The scene is illuminated with a spherical emitter that simulates a bulb with power value of 1000W and efficacy of 17,6 lm/W.  The overall setup is simple and designed to be reproducible with real objects and material samples.

	\paragraph*{Training} We use the synthetic data to train our neural networks. In particular, only the views at 30$^{\circ}$ and 90$^{\circ}$ with the color checker are used for training. We found that these views are sufficiently representative of most materials and we empirically proved that they are optimal for this task. The 12 other views are used at test time to validate the generalization of the predicted material to novel views (see supplementary material for an example). We separate 2/3 of the data for the training, 1/6 for validation and 1/6 for test. For some of the models, We perform data augmentation by adding Gaussian noise with a mean of 0 and random standard deviation between 0 and 4. We evaluate the performance of this operation in the following section. 
	
	We first train the networks \CNNo and \CNNw using the parameter loss of Equation~\ref{eq:ploss} with Mini-batch Stochastic Gradient Descent, learning rate of 0.003 and random initialization of the weights. Then, we use the features learned in this first step to train the nested module. For this aim, the feature vectors are kept fixed (the weights of the convolutionals networks aren't updated) and the weights of the new fully-connected layers in the FC networks are randomly initialized and trained with a learning rate of 0.03.

	\subsection{Quantitative Evaluation}
	We quantitatively evaluate our method using the test set of our MaxBRDF dataset, from which we know both the ground truth parameters as well as the rendered images of each sample.

	\begin{figure}
		\centering
		\includegraphics[width=0.9\columnwidth]{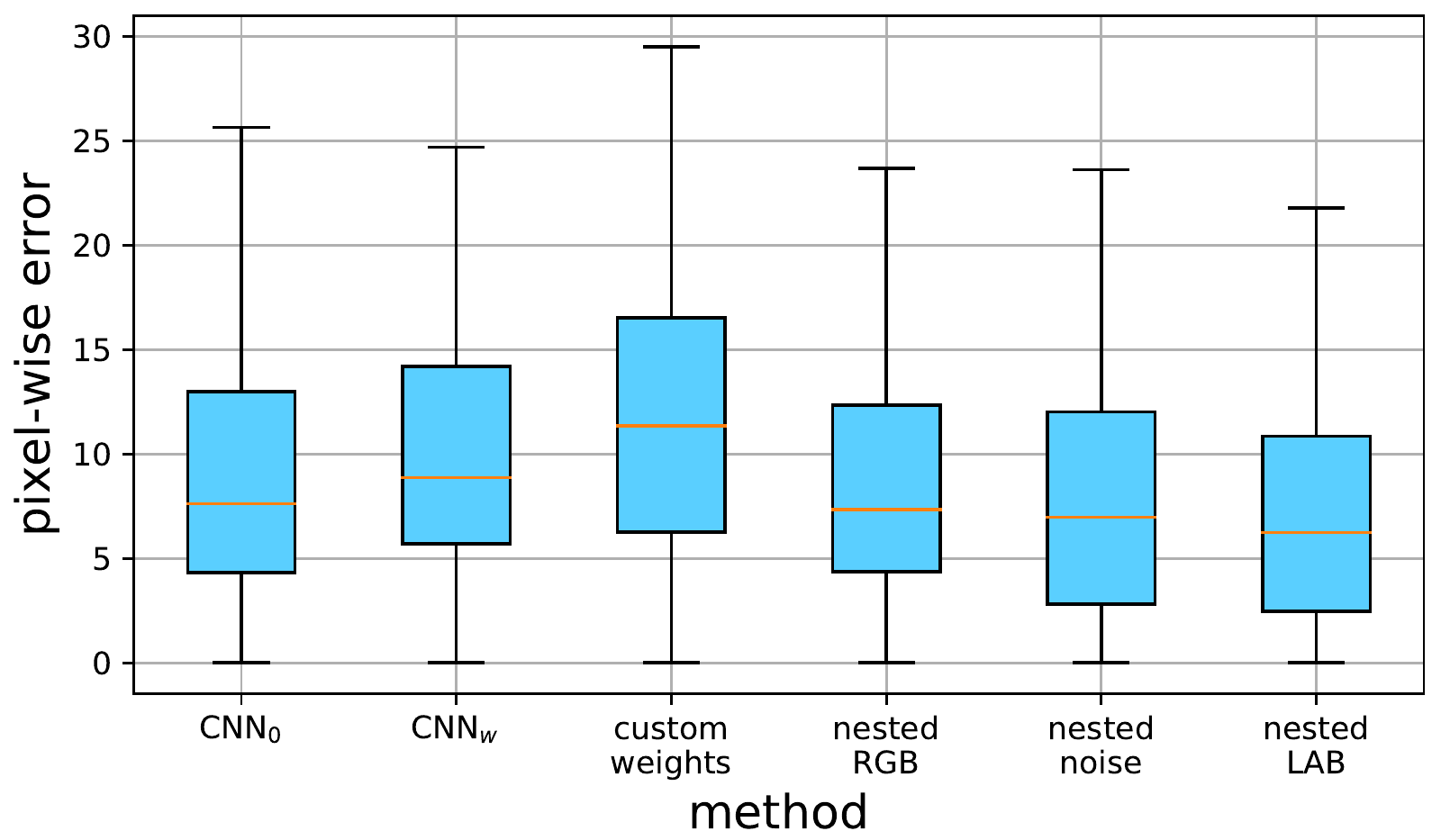}
		\caption{Pixel-wise error of the predictions on the testset of our MaxBRDF dataset. Each box depicts the results obtained using a different method. Nested LAB, described in Section \ref{sec:regression}, outperforms the rest.}
		\label{fig:pixelwise_synthetic}
	\end{figure}
	\begin{figure}
		\centering
		\begin{subfigure}[t]{0.24\columnwidth}
			\includegraphics[width=\textwidth]{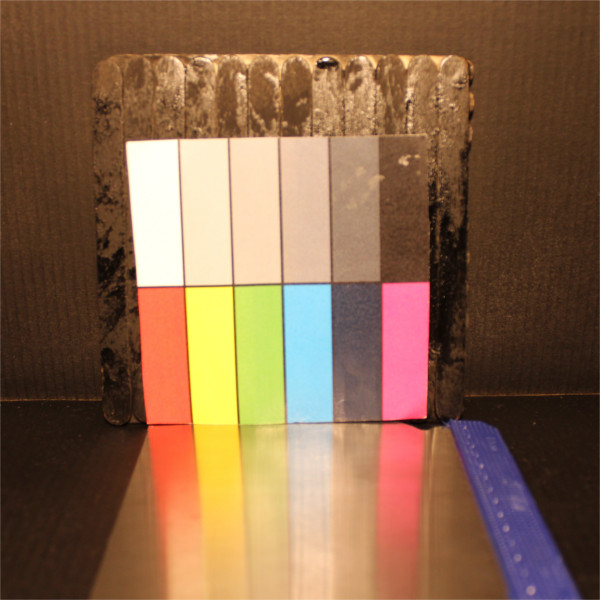}
			\caption*{\scriptsize{Captured 30$^{\circ}$}}
		\end{subfigure}
		\begin{subfigure}[t]{0.24\columnwidth}
			\includegraphics[width=\textwidth]{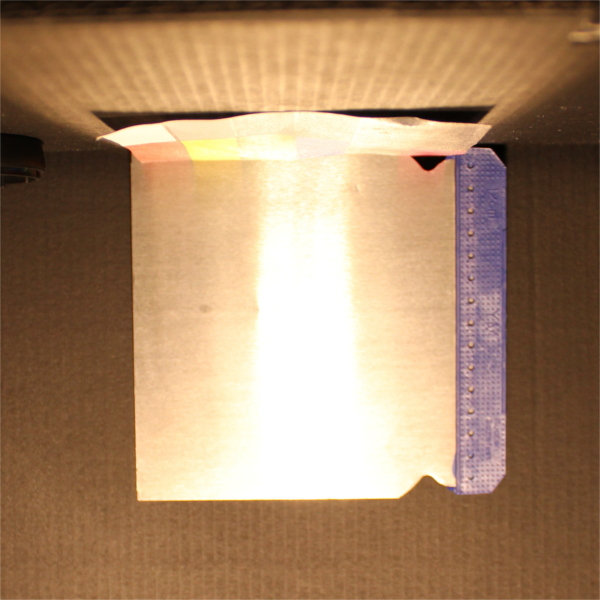}
			\caption*{\scriptsize{Captured 90$^{\circ}$}}
		\end{subfigure}
		\begin{subfigure}[t]{0.24\columnwidth}
			\includegraphics[width=\textwidth]{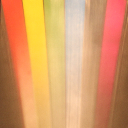}
			\caption*{\scriptsize{Homographied 30$^{\circ}$}}
		\end{subfigure}
		\begin{subfigure}[t]{0.24\columnwidth}
			\includegraphics[width=\textwidth]{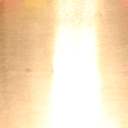}
			\caption*{\scriptsize
				{Homographied 90$^{\circ}$}}
		\end{subfigure}
		\caption{Sample of our test set of real materials. Using a low cost setup, we capture them from two viewpoints.}
		\label{fig:capture_setup}
	\end{figure}
	
	Figure \ref{fig:pixelwise_synthetic} depicts the pixel-wise mean square error (MSE) of the rendered predictions and demonstrates that our approach (``nested LAB'') outperforms the baseline and other intermediate solutions.
	In particular, we show MSE per-pixel error for:
	first, our $\text{CNN}_0$ and $\text{CNN}_\text{w}$ alone, described in Section \ref{sec:feature_extraction}. As expected, these two networks are good at extracting discriminative features of the input images, but struggle to provide accurate predictions because they have to implicitly learn parameter dependencies;
	we also experimented with assigning explicit weights (in Figure \ref{fig:pixelwise_synthetic}, ``custom weights'') to each of the parameters to the loss function $\mathcal{L}_{total}$ defined in Equation \ref{eq:global_loss}, hoping that these weights would ease the learning process. However, results were unsatisfactory, probably due to the complex parameter dependency of our BRDF model; finally, we show results with our nested approach using different representations to compute the error of the reflectance parameters $\RGBN$ and $\RGBZ$ (in Figure \ref{fig:pixelwise_synthetic}, ``nested RGB'', ``nested noise'' and ``nested LAB'').
	We conclude that our nested approach using the LAB color space to encode $\RGBN$ and $\RGBZ$, which includes Gaussian noise augmentation in the training set, outperforms any other method.



	{
		\newcommand{\figScale}{0.18\textwidth}
		\begin{figure*}[t]
			\centering
			\resizebox{\textwidth}{!}{
				\begin{tikzpicture}
				\node[rotate=90] (textA) at (0,0) {\Large{Real photo}};
				\node[right of=textA, xshift=1.1cm] (2d_1) 
				{\includegraphics[width=\figScale]{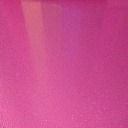}};
				\node[right of=2d_1, xshift=2.3cm] (2d_2) 
				{\includegraphics[width=\figScale]{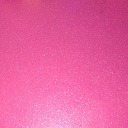}};
				\node[right of=2d_2, xshift=2.7cm] (2d_3) 
				{\includegraphics[width=\figScale]{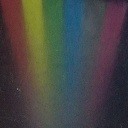}};
				\node[right of=2d_3, xshift=2.3cm] (2d_4) 
				{\includegraphics[width=\figScale]{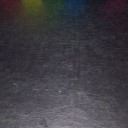}};
				\node[right of=2d_4, xshift=2.7cm] (2d_5) 
				{\includegraphics[width=\figScale]{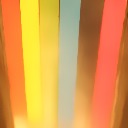}};
				\node[right of=2d_5, xshift=2.3cm] (2d_6) 
				{\includegraphics[width=\figScale]{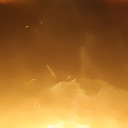}};
				\node[right of=2d_6, xshift=2.7cm] (2d_7) 
				{\includegraphics[width=\figScale]{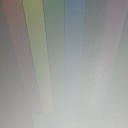}};
				\node[right of=2d_7, xshift=2.3cm] (2d_8) 
				{\includegraphics[width=\figScale]{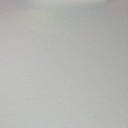}};
				\node[right of=2d_8, xshift=2.7cm] (2d_9) 
				{\includegraphics[width=\figScale]{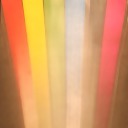}};
				\node[right of=2d_9, xshift=2.3cm] (2d_10) 
				{\includegraphics[width=\figScale]{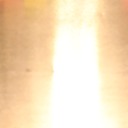}};
				%
				\node[rotate=90, left of=textA, xshift=-2.3cm] (textB)  {\Large{Prediction}};
				\node[right of=textB, xshift=1.1cm] (3d_1) 
				{\includegraphics[width=\figScale]{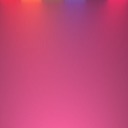}};
				\node[right of=3d_1, xshift=2.3cm] (3d_2) 
				{\includegraphics[width=\figScale]{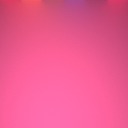}};
				\node[right of=3d_2, xshift=2.7cm] (3d_3) 
				{\includegraphics[width=\figScale]{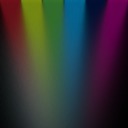}};
				\node[right of=3d_3, xshift=2.3cm] (3d_4) 
				{\includegraphics[width=\figScale]{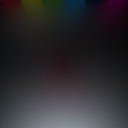}};
				\node[right of=3d_4, xshift=2.7cm] (3d_5) 
				{\includegraphics[width=\figScale]{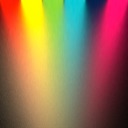}};
				\node[right of=3d_5, xshift=2.3cm] (3d_6) 
				{\includegraphics[width=\figScale]{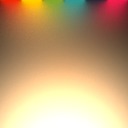}};
				\node[right of=3d_6, xshift=2.7cm] (3d_7) 
				{\includegraphics[width=\figScale]{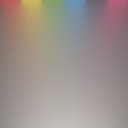}};
				\node[right of=3d_7, xshift=2.3cm] (3d_8) 
				{\includegraphics[width=\figScale]{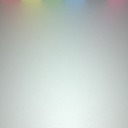}};
				\node[right of=3d_8, xshift=2.7cm] (3d_9) 
				{\includegraphics[width=\figScale]{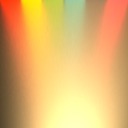}};
				\node[right of=3d_9, xshift=2.3cm] (3d_10) 
				{\includegraphics[width=\figScale]{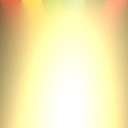}};
				%
				\node[rotate=90, left of=textB, xshift=-2.7cm] (textC)  {\Large{Real photo}};
				\node[right of=textC, xshift=1.1cm] (4d_1) 
				{\includegraphics[width=\figScale]{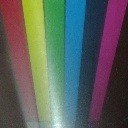}};
				\node[right of=4d_1, xshift=2.3cm] (4d_2) 
				{\includegraphics[width=\figScale]{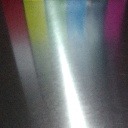}};
				\node[right of=4d_2, xshift=2.7cm] (4d_3) 
				{\includegraphics[width=\figScale]{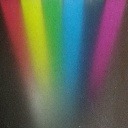}};
				\node[right of=4d_3, xshift=2.3cm] (4d_4) 
				{\includegraphics[width=\figScale]{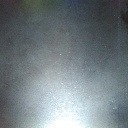}};
				\node[right of=4d_4, xshift=2.7cm] (4d_5) 
				{\includegraphics[width=\figScale]{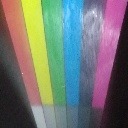}};
				\node[right of=4d_5, xshift=2.3cm] (4d_6) 
				{\includegraphics[width=\figScale]{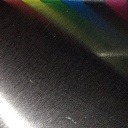}};
				\node[right of=4d_6, xshift=2.7cm] (4d_7) 
				{\includegraphics[width=\figScale]{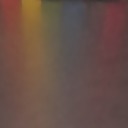}};
				\node[right of=4d_7, xshift=2.3cm] (4d_8) 
				{\includegraphics[width=\figScale]{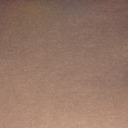}};
				\node[right of=4d_8, xshift=2.7cm] (4d_9) 
				{\includegraphics[width=\figScale]{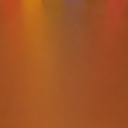}};
				\node[right of=4d_9, xshift=2.3cm] (4d_10) 
				{\includegraphics[width=\figScale]{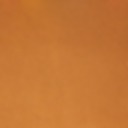}};
				%
				\node[rotate=90, left of=textC, xshift=-2.3cm] (textD)  {\Large{Prediction}};
				\node[right of=textD, xshift=1.1cm] (5d_1) 
				{\includegraphics[width=\figScale]{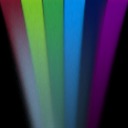}};
				\node[right of=5d_1, xshift=2.3cm] (5d_2) 
				{\includegraphics[width=\figScale]{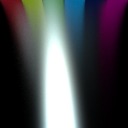}};
				\node[right of=5d_2, xshift=2.7cm] (5d_3) 
				{\includegraphics[width=\figScale]{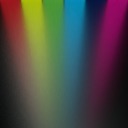}};
				\node[right of=5d_3, xshift=2.3cm] (5d_4) 
				{\includegraphics[width=\figScale]{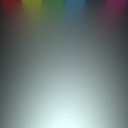}};
				\node[right of=5d_4, xshift=2.7cm] (5d_5) 
				{\includegraphics[width=\figScale]{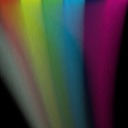}};
				\node[right of=5d_5, xshift=2.3cm] (5d_6) 
				{\includegraphics[width=\figScale]{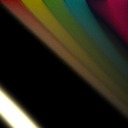}};
				\node[right of=5d_6, xshift=2.7cm] (5d_7) 
				{\includegraphics[width=\figScale]{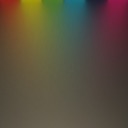}};
				\node[right of=5d_7, xshift=2.3cm] (5d_8) 
				{\includegraphics[width=\figScale]{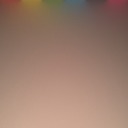}};
				\node[right of=5d_8, xshift=2.7cm] (5d_9) 
				{\includegraphics[width=\figScale]{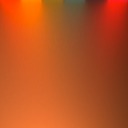}};
				\node[right of=5d_9, xshift=2.3cm] (5d_10) 
				{\includegraphics[width=\figScale]{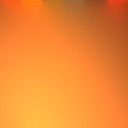}};
				%
				\node[rotate=90, left of=textD, xshift=-2.7cm] (textE)  {\Large{Real photo}};
				\node[right of=textE, xshift=1.1cm] (6d_1) 
				{\includegraphics[width=\figScale]{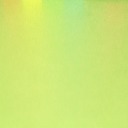}};
				\node[right of=6d_1, xshift=2.3cm] (6d_2) 
				{\includegraphics[width=\figScale]{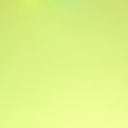}};
				\node[right of=6d_2, xshift=2.7cm] (6d_3) 
				{\includegraphics[width=\figScale]{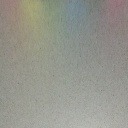}};
				\node[right of=6d_3, xshift=2.3cm] (6d_4) 
				{\includegraphics[width=\figScale]{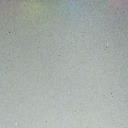}};
				\node[right of=6d_4, xshift=2.7cm] (6d_5) 
				{\includegraphics[width=\figScale]{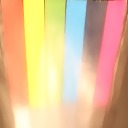}};
				\node[right of=6d_5, xshift=2.3cm] (6d_6) 
				{\includegraphics[width=\figScale]{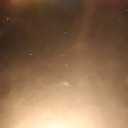}};
				\node[right of=6d_6, xshift=2.7cm] (6d_7) 
				{\includegraphics[width=\figScale]{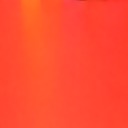}};
				\node[right of=6d_7, xshift=2.3cm] (6d_8) 
				{\includegraphics[width=\figScale]{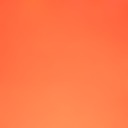}};                
				\node[right of=6d_8, xshift=2.7cm] (6d_9) 
				{\includegraphics[width=\figScale]{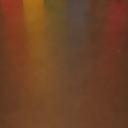}};
				\node[right of=6d_9, xshift=2.3cm] (6d_10) 
				{\includegraphics[width=\figScale]{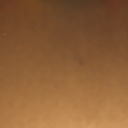}};        
				\node[rotate=90, left of=textE, xshift=-2.3cm] (textF)  {\Large{Prediction}};
				\node[right of=textF, xshift=1.1cm] (7d_1) 
				{\includegraphics[width=\figScale]{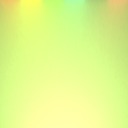}};
				\node[right of=7d_1, xshift=2.3cm] (7d_2) 
				{\includegraphics[width=\figScale]{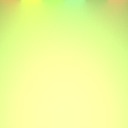}};
				\node[right of=7d_2, xshift=2.7cm] (7d_3) 
				{\includegraphics[width=\figScale]{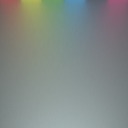}};
				\node[right of=7d_3, xshift=2.3cm] (7d_4) 
				{\includegraphics[width=\figScale]{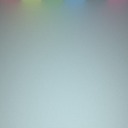}};
				\node[right of=7d_4, xshift=2.7cm] (7d_5) 
				{\includegraphics[width=\figScale]{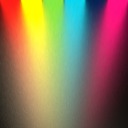}};
				\node[right of=7d_5, xshift=2.3cm] (7d_6) 
				{\includegraphics[width=\figScale]{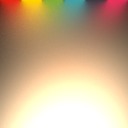}};
				\node[right of=7d_6, xshift=2.7cm] (7d_7) 
				{\includegraphics[width=\figScale]{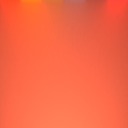}};
				\node[right of=7d_7, xshift=2.3cm] (7d_8) 
				{\includegraphics[width=\figScale]{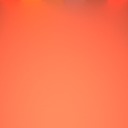}};
				\node[right of=7d_8, xshift=2.7cm] (7d_9) 
				{\includegraphics[width=\figScale]{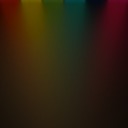}};
				\node[right of=7d_9, xshift=2.3cm] (7d_10) 
				{\includegraphics[width=\figScale]{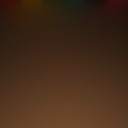}};    
				\end{tikzpicture}
			}
			\caption{Qualitative evaluation on real materials. Each 4x4 block shows a pair of homographied images of a real material (captured at 30$^{\circ}$ and 90$^{\circ}$), and the predicted material rendered in the same configuration. This results demonstrates that the proposed method is capable of capturing complex appearance behaviors such as strong highlights and anisotropies.}
			\label{fig:qualitative_real}
		\end{figure*}
	}
	\subsection{Qualitative Evaluation}
	\label{sec:qualitative_evaluation}
	
	The images of materials predicted by our method using the test set of MaxBRDF dataset are really accurate. Several images on how it generalizes to new viewpoints can be found in the supplementary material. However, our final goal is to create a method that estimates the parameters of pictures of real materials. 
	
	Even though our new MaxBRDF dataset features a large variety of physically-based rendered materials, it does not contain any \textit{real} image to use for testing.
	
	We overcome this limitation by capturing a new test set featuring real-world materials in a low-cost setup, as shown in Figure \ref{fig:capture_setup}. Capturing the ground truth properties of those materials is a challenging task, even with the help of a goniorreflectometer and BRDF fitting methods which are beyond the scope of this work. However, we can use these images to visually assess the quality of the predictions and evaluate the potential of our method for casual BRDF acquisition on the wild.
	
	For each material, we take a photo with a handheld camera approximately at 30$^{\circ}$ and 90$^{\circ}$ in a controlled environment, where a unique light is located roughly at the same position as in the synthetic dataset. As shown in Figure \ref{fig:capture_setup}, following our pre-processing step, each of these image pairs is homographied before being fed into our network.
	
	Figure \ref{fig:qualitative_real} shows 15 pairs of homographied real images of complex materials, and the corresponding renders of the predicted BRDF. We demonstrate that our method obtains highly accurate results, even for materials featuring complex anisotropic and specular behaviors.
	
	Furthermore, in the supplementary video, we show highly realistic animations of the predicted BRDF of real materials, using a moving camera. This demonstrates that our method can be used to render novel views of complex materials captured with just two quick shots. 
	
	%
	%
\section{Conclusions}
In this paper, we have presented the first method to estimate the complex appearance of anisotropic materials using a novel architecture based on nested neural networks. We have shown that given only two views of a material sample at 30$^{\circ}$ and 90$^{\circ}$ degrees, our model can extrapolate the estimations to the remaining fourteen views of our dataset. We have shown the performance of our method with quantitative and qualitative experiments on synthetic and real data as input, providing compelling results in both scenarios. We have also presented the first method to achieve non-convex parameter fitting by using a hybrid approach between feature learning with convolutional neural networks and classification-based regression using a nested fully connected architecture. We believe we are the first to provide such kind of solution to the problem outperforming end-to-end baseline architectures.


{\small
\bibliographystyle{ieee}
\bibliography{BSDF_estimation}
}

\end{document}